\def\year{2017}\relax
\documentclass[letterpaper]{article}
\usepackage{aaai17}
\usepackage{times}
\usepackage{helvet}
\usepackage{amsfonts}
\usepackage{courier}
\usepackage{amsmath}
\usepackage{amssymb}
\usepackage{hyperref}
\usepackage{algorithm}
\usepackage{algorithmic}
\usepackage{caption}
\usepackage{subfloat}
\usepackage{graphicx}
\usepackage{float}
\usepackage{bm}
\usepackage{subfig}

\newcommand\blfootnote[1]{%
  \begingroup
  \renewcommand\thefootnote{}\footnote{#1}%
  \addtocounter{footnote}{-1}%
  \endgroup
}

\begin{document}
%
\title{Option Discovery in Hierarchical Reinforcement Learning using Spatio-Temporal Clustering}
\author{
Aravind Lakshminarayanan$^{*}$ \\ Indian Institute of Technology, Madras \\ aravindsrinivas@gmail.com
\And Ramnandan Krishnamurthy$^{*,\dagger}$ \\ Microsoft Corporation, Redmond \\ nandparikrish@gmail.com
\AND Peeyush Kumar$^{*, \dagger}$ \\ University of Washington, Seattle \\ agoovi@gmail.com
\And Balaraman Ravindran \\ Indian Institute of Technology, Madras \\ ravi@cse.iitm.ac.in}

\maketitle

\blfootnote{
\begin{itemize}
    \item $*$ - Equal Contribution
    \item $\dagger$ - Work done while authors were at IIT Madras

\end{itemize}
}

\begin{abstract}
This paper introduces an automated skill acquisition framework in reinforcement learning which involves identifying a hierarchical description of the given task in terms of abstract states and extended actions between abstract states. Identifying such structures present in the task provides ways to simplify and speed up reinforcement learning algorithms. These structures also help to generalize such algorithms over multiple tasks without relearning policies from scratch. We use ideas from dynamical systems to find metastable regions in the state space and associate them with abstract states. The spectral clustering algorithm PCCA+ is used to identify suitable abstractions aligned to the underlying structure. Skills are defined in terms of the sequence of actions that lead to transitions between such abstract states. The connectivity information from PCCA+ is used to generate these skills or options. These skills are independent of the learning task and can be efficiently reused across a variety of tasks defined over the same model. This approach works well even without the exact model of the environment by using sample trajectories to construct an approximate estimate. We also present our approach to scaling the skill acquisition framework to complex tasks with large state spaces for which we perform state aggregation using the representation learned from an action conditional video prediction network and use the skill acquisition framework on the aggregated state space.
\end{abstract}

\section{Introduction}

The core idea of hierarchical reinforcement learning is to break down the reinforcement learning problem into subtasks through a hierarchy of abstractions. In terms of Markov Decision Processes (MDP), a well studied framework in reinforcement learning literature \cite{rlbook}, one way of looking at the full reinforcement learning problem is to assume that the agent is in one state of the MDP at each time step. The agent then performs one of several possible primitive actions which along with the current state decides the next state. However, for large problems, this can lead to too much granularity: when the agent has to decide on each and every primitive action at every granular state, it can often lose sight of the {\it bigger} picture \cite{finney2002thing}. If a series of actions can be abstracted out as a single {\it macro} action, the agent can just remember the series of actions that was useful in getting it to a temporally distant useful state from the initial state. This is typically referred to as a skill and more specifically, as an option in \cite{sutton1999between}. A good analogy is a human planning his movement from current location $A$ to destination $B$. We identify intermediate destinations $C_i$ to lead us from $A$ to $B$ when planning from $A$ rather than worrying about the exact mechanisms of immediate movement at $A$ which are {\it abstracted over}. Options are a convenient way to formalise this abstraction. In keeping with the general philosophy of reinforcement learning, we want to build agents that can automatically discover options without prior knowledge, purely by exploring the environment. Thus, our approach falls into the broad category of {\it automated discovery of skills}. Such skills which are learnt in one task could be easily reused in a different task if necessary. 

Our focus in this paper is to present our framework on automated discovery of skills. Automated discovery of skills or options has been an active area of research and several approaches have been proposed for the same. The current methods could be broadly classified into sample trajectory based and partition based methods. Some of them are:

\begin{itemize}
    \item Identifying bottlenecks in the state space, where the state space is partitioned into sets. The transitions between two sets of states that are rare introduce {\it bottleneck} states at the respective points of such transitions. Policies to reach such bottleneck states are cached as options - for example, \cite{mcgovern2001automatic}.
    \item Using the structure present in a factored state representation to identify sequences of actions that cause what are otherwise infrequent changes in the state variables: these sequences are cached away as options \cite{hengst2004model}.
    \item Obtaining a graphical representation of an agent's interaction with its environment and using betweenness centrality measures to identify subtasks \cite{csimcsek2009skill}.
    \item Using clustering methods (spectral or otherwise) to separate out different strongly connected components of the MDP and identifying {\it bottlenecks} that connect different clusters \cite{menache2002q}.
    
\end{itemize}

There are certain deficiencies with the above methods, even though they have had varying degrees of success. Bottleneck based approaches do not have a natural way of identifying the part of the state space where options are applicable without external knowledge about the problem domain. Spectral methods need some form of regularization in order to prevent unequal splits that might lead to arbitrary splitting of the state space. There have also been attempts to learn skill acquisition in robotics. \cite{konidaris2011autonomous} attempt to discover skills for a robot given abstractions of the surroundings it has to operate with. \cite{ranchod2015nonparametric} try to use inverse reinforcement learning for skill discovery with reward segmentation. However, the above two works assume the knowledge of the abstractions of the world and demonstrations of expert trajectories respectively while our goal is to acquire skills and discover the abstractions without any prior knowledge. 

We present a framework that detects well-connected or meta-stable regions of the state space from an MDP model estimated from trajectories. We use PCCA+, a spectral clustering algorithm from conformal dynamics \cite{weber2004perron} that not only partitions the MDP into different regions but also returns the connectivity information between the regions, unlike other clustering approaches used earlier for option discovery. This helps us to build an abstraction of the MDP, where we call the regions identified by PCCA+ as {\it abstract states}. Then we define options that take an agent from one abstract state to another connected abstract state. Since PCCA+ also returns the membership function between states and abstract states, we propose a very efficient way of constructing option policies directly from the membership function, which is to perform {\it hill climbing} at the granular states on the membership function of the {\it destination} abstract state, to yield the option policy without further learning. Since the abstract states are aligned with the underlying structure of the MDP the same option policies could be efficiently reused across multiple tasks in the same MDP. Once we have these options, we could use standard reinforcement learning algorithms to learn a policy over these subtasks to solve a given task. Specifically, we use SMDP Q-Learning \cite{sutton1999between} and Intra Option Q-Learning \cite{sutton1998intra} for our experiments. Note that our approach works well even without the exact model of the full MDP, by being able to work on approximate estimate of the model using sample trajectories, as demonstrated by the experiments.

Finally, we present our attempt to extend our pipeline to large state spaces where spectral methods on the original state space are infeasible. We therefore propose to operate the PCCA+ pipeline on an {\it aggregated state space}. The original state space is {\it aggregated} through clustering on the representation of the state space learned using deep neural networks. Specifically, we try to learn a representation that captures spatio-temporal structure of the state-space, for which we borrow the framework of \cite{oh2015action} which uses a Convolutional-LSTM Action Conditional Video Prediction Network to predict the next frames of the game conditioned on the trajectory so far. We use the Arcade Learning Environment (ALE) platform \cite{Bellemare} which provides a simulator for Atari 2600 games. We present our results on the Atari 2600 game Seaquest which is a relatively complex game and requires abstract moves like {\it filling up oxygen, evading bullets, shooting enemies}.

We summarize our key contributions in this paper below:

\begin{itemize}
    \item A novel automated skill acquisition pipeline to operate without prior knowledge or abstractions of the world. The pipeline uses a spectral clustering algorithm from conformal dynamics - PCCA+, which builds an abstraction of the MDP by segmenting the MDP into different regions called {\it abstract states} and also provides the membership function between the MDP states and {\it abstract states}.
    \item An elegant way of composing options (sequence of actions to move from one abstract state to another) from the membership function so obtained by doing hill climbing on the membership function of the {\it target abstract state} at the granular states belonging to the {\it current abstract state}.
    \item Extension of the pipeline to complex tasks with large state spaces, for which we propose to run the pipeline on an {\it aggregated state space} due to infeasibility to operate on the original state space of the MDP. To aggregate the state space, we learn a representation capturing spatio-temporal aspects of the state space using a deep action-conditional video prediction network.
\end{itemize}

Our approach can be considered as an attempt to address automated option discovery in RL using spatio-temporal clustering: Spatial because the abstract states are segmenting the state space into abstractions (clusters); and Temporal since the transition structure of the MDP is used to discover these abstractions.


\section{Preliminaries}
\subsection{Markov Decision Process}

Markov Decision Process (MDP) is a widely used framework on which reinforcement learning algorithms are used to learn control policies. A finite discrete MDP is formalised as $M=(S,A,P,R)$ with $S$ with a finite discrete state space $S$, finite action space $A$, a distribution over the next states $P(s,a,s')$ when an action $a$ is performed at state $s$, and a corresponding reward $R(s,a,s')$ specifying a scalar cost or reward associated with that transition. A policy is a sequence of actions involved in performing a task on the MDP, while value functions are a measure of the expected long term return in following a policy in the MDP.
\subsection{Options}

The option framework is one of the formalisations used to represent hierarchies in RL \cite{sutton1999between}. Formally, an option is a tuple $O=(I,\mu,\beta)$ here:
\begin{itemize}
    \item $I$ is the initiation set: a set of states from which the action can be activated
    \item $\mu$ is a policy function where $\mu(s,a)$ represents the preference value given to action $a$ in state $s$ following option $O$.
    \item $\beta$ is the termination function: When an agent enters a state $s$ while following option $O$, the option could be terminated with a probability $\beta(s)$.
\end{itemize}

When the agent is in state $s$ where it can start an option, it can choose from all the options $O$ for which $s\in I(O)$ and all primitive actions that can be taken in $s$. This choice is dictated by the policy guiding the agent. 

\subsection{SMDP Value Learning}

An MDP appended with a set of options is a Semi-Markov Decision Process (SMDP). In an SMDP, at each time step, the agent selects an option which can be initiated at that state and follows the corresponding option policy until termination. In SMDP theory, the effects of an option are modelled using $r_{s}^o$ and $p_{ss^{'}}^o$ which represent the total return and the probability of terminating at state $s^{'}$ for an option $o$ initiated at state $s$. Each option is viewed as an indivisible unit and its structure not utilized. The update rule of SMDP Q-learning~\cite{sutton1999between} is given by
$$Q(s,o)\leftarrow Q(s,o)+\alpha[r+\gamma^k max_{a\in O}Q(s,a)-Q(s,o)]$$ where $Q$ is the action/option value function.



\subsection{Intra-Option Value Learning} 

A major drawback of SMDP learning methods is that the option needs to be completely executed upto termination before learning about its outcome. At any point of time, we would potentially be learning about the value function of only option or a primitive action. This slows down learning significantly for large problems of the scale of Atari 2600 games. \cite{sutton1998intra} propose to use all options that are consistent with an observed trajectory and have their values updated efficiently. This can help learn the values of certain options without even executing those options. Let $o$ be an option chosen at state $s_t$ at time $t$ which terminates at time $t+\tau$. Instead of using a single training example to update $Q(s_t,o)$ as in SMDP methods, the Markov nature of the option $o$ is exploited and each of the transitions in the $\tau$ time steps are valid training examples. The Q-value of options $o'$ consistent with actions taken during those time steps are also updated allowing for efficient off-policy learning.

\subsection{Perron Cluster Analysis (PCCA+)}

Given an algebraic representation of the graph representing an MDP, we want to find suitable abstractions aligned to the underlying structure. We use a spectral clustering algorithm to do this. Central to the idea of spectral clustering is the graph Laplacian which is obtained from the similarity graph. In this approach, the spectra of the Laplacian $\mathcal{L}$ (derived from the adjacency matrix $\mathcal{S}$) is constructed and the best transformation of the spectra is found such that the transformed basis aligns itself with the clusters of data points in the eigenspace. A projection method described in \cite{weber2004perron} is used to find the membership of each of the states to a set of $k$ special points lying on the transformed basis, which are identified as vertices of a simplex in the $\mathbb{R}^k$ subspace (the Spectral Gap method is used to estimate the number of clusters $k$). For the first order perturbation, the simplex is just a linear transformation around the origin and to find the simplex vertices, one needs to find the $k$ points which form a convex hull such that the deviation of all the points from this hull is minimized. This is achieved by finding the data point which is located farthest from the origin and iteratively identify data points which are located farthest from the hyperplane fit to the current set of vertices. Refer to Algorithm 1 for details.

 \begin{figure}[H]
   \begin{center}
     \includegraphics [scale=0.27]{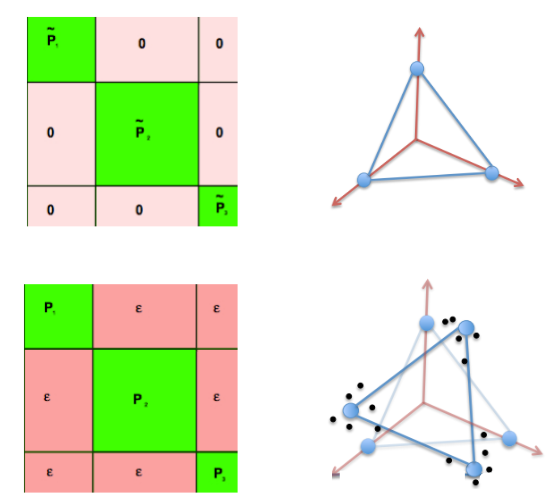}
     \caption {Simplex First order and Higher order Perturbation: Shows the visualization of the first order perturbation assumption to identify the simplex vertices on data points which have higher order perturbations. The first case shows data points which satisfy the first order assumption and hence the simplex fits perfectly without any noise. In the second case, the simplex vertices obtained through the linear transformation are only able to capture the structure with some noise and these deviations could be understood as the effect of the higher order perturbations present.}
   \label{fig:simplex}
   \end{center}
 \end{figure}

\begin{algorithm}[H] \label{algorithm}
\caption{PCCA+}
\begin{algorithmic}[1]
\STATE Construct Laplacian $\mathcal{L}$ 
\STATE Compute $n$ (number of vertices) eigenvalues of $\mathcal{L}$ in descending order 
\STATE Choose first $k$ eigenvalues for which $\frac{e_k-e_{k+1}}{1-e_{k+1}}>t_c$ (Spectral Gap Threshold). 
\STATE Compute the eigenvectors for corresponding eigenvalues ($e_1,e_2,\cdots,e_k$) and stack them as column vectors in eigenvector matrix $\mathcal{Y}$. 
\STATE Let's denote the rows of $\mathcal{Y}$ as $\mathcal{Y}(1),\mathcal{Y}(2),\cdots,\mathcal{Y}(N) \in \mathbb{R}^k$.  
\STATE Define $\pi$(1) as that index, for which $||Y(\pi(1))||_2$ is maximal. Define $\gamma_1=span\{Y(\pi(1))\}$.
\STATE \textbf{For} $i = 2,\cdots,k$: Define $\pi_i$ as that index, for which the distance to the hyperplane $\gamma_{i-1}$, i.e., $||\mathcal{Y}(\pi_i)-\gamma_{i-1}||_2$ is maximal. Define $\gamma_i=span\{\mathcal{Y}(\pi_1),\cdots,\mathcal{Y}(\pi_i)\}$. To compute $||\mathcal{Y}(\pi_i)-\gamma_{i-1}||_2$, use $||\mathcal{Y}(\pi_i)-\gamma_{i-1}||_2=||\mathcal{Y}(\pi_i)-\gamma_{i-1}^{T}((\gamma_{i-1}\gamma_{i-1}^{T})^{-1}\gamma_{i-1})\mathcal{Y}(\pi_i)^T)||$ 
\end{algorithmic}
\end{algorithm}

\begin{figure} 
\centering

\subfloat[PCCA+ vs rest - Room-in-a-room Domain]{
        \label{fig:chain_w}
        \includegraphics[width=0.25\textwidth]{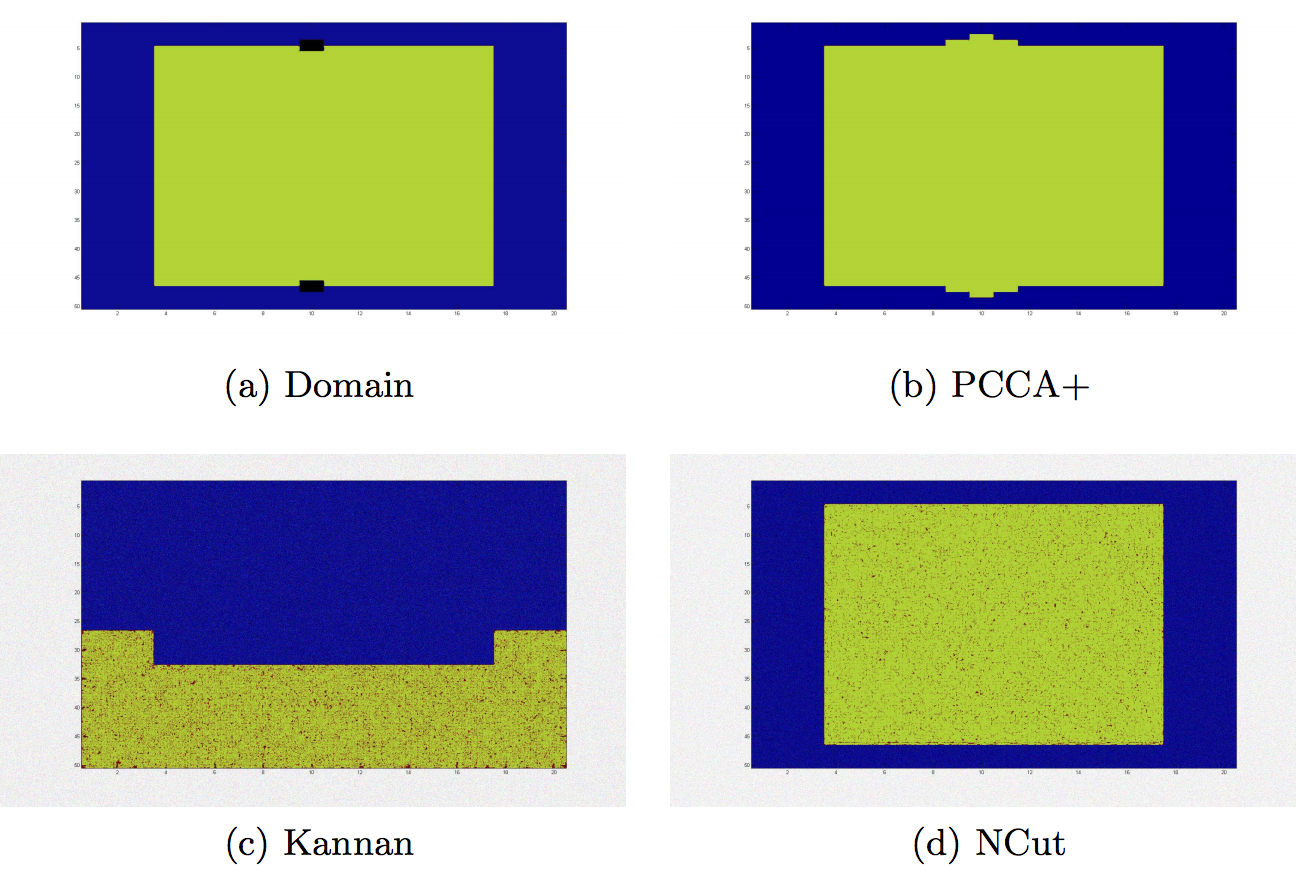}}
\subfloat[PCCA+ vs rest - Maze Domain]{
        \label{fig:w1}
        \includegraphics[width=0.25\textwidth]{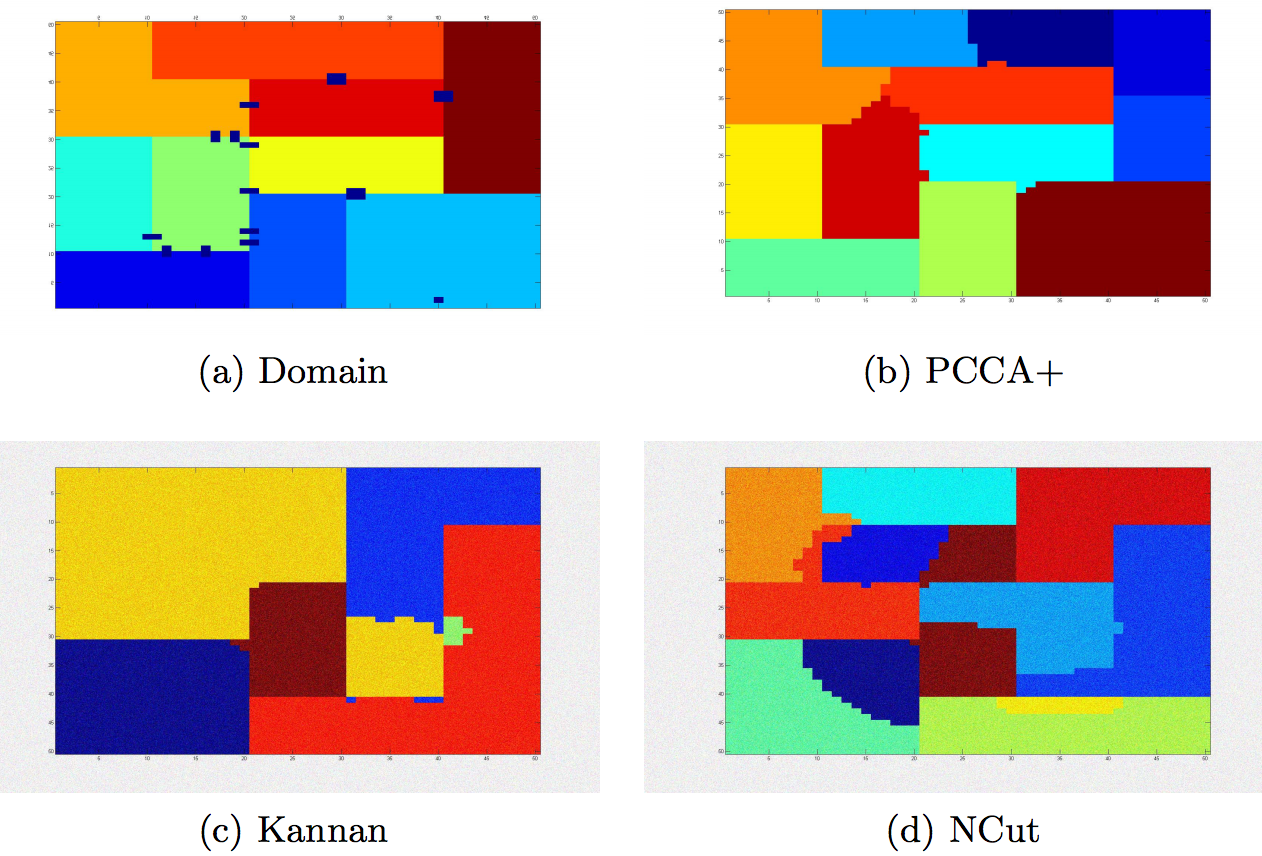} } \\
%
%
\subfloat[PCCA+ vs NCut - Real Image]{
\label{fig:w3}
       \includegraphics[width=0.25\textwidth]{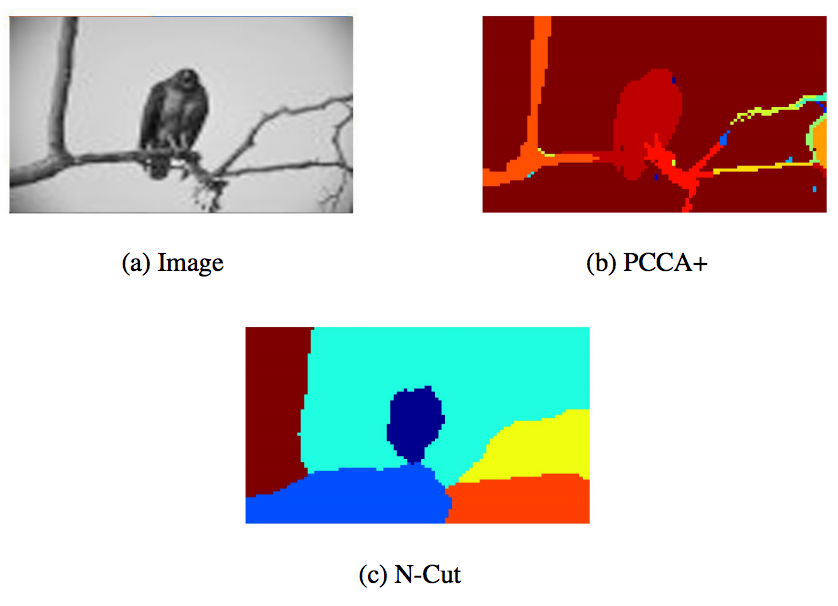}}
\subfloat[PCCA+ Membership functions visualization]{
        \label{fig:w4}
        \includegraphics[width=0.25\textwidth]{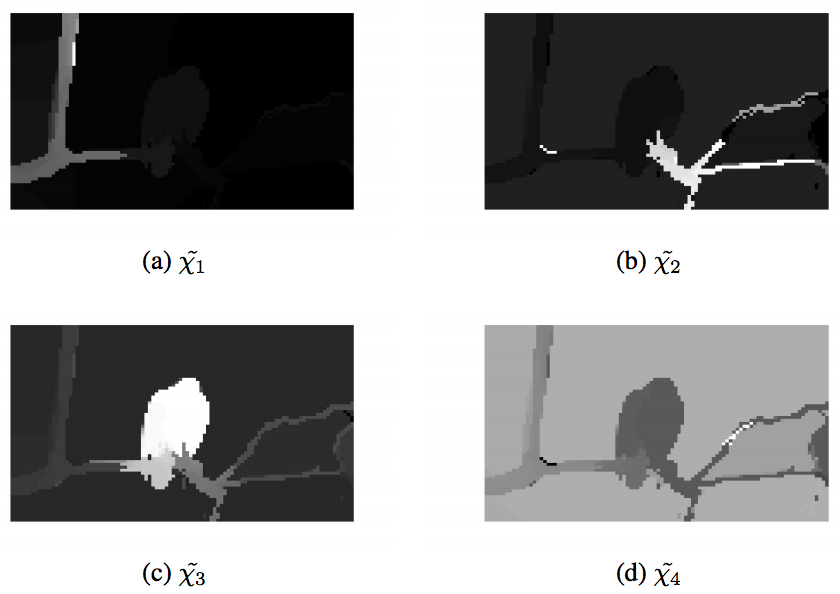}}

\caption{PCCA+ visualizations}
\label{fig:kohler}
\end{figure}

\subsubsection{Comparison to other clustering algorithms}
We are in particular concerned with dynamical systems where there are no a priori assumptions on the size and the relative placements of the metastable states. So it becomes essential for the clustering algorithm to work in a generalized setting as far as possible. We tested different spectral clustering algorithms on different classes of problems. PCCA+ was found to give better results (Fig \ref{fig:kohler}) in capturing the structure (topology). Normalized Cut (NCut) Algorithms by \cite{shi2000normalized},\cite{ng2002spectral} and PCCA+ \cite{weber2004perron} have deep similarities, but the main difference behind the usage of PCCA+ is the change of point of view for identification of metastable states from crisp clustering to that of relaxed almost invariant sets. Both the methods are similar till the identification of the eigenvector matrix  as points in $n$ dimensional subspace but after this, \cite{shi2000normalized},\cite{ng2002spectral} use standard clustering algorithms like K-Means while PCCA+ uses a mapping to the simplex structure. Since the bottlenecks occur rather rarely, it is shown by \cite{weber2004perron} that in general for such cases, soft methods outperform crisp clustering. Although the other spectral algorithms result in good clusters for the simple room-in-a-room domain, they give poor results on more topologically complex spaces and need some form of regularization to work well in such settings (Fig \ref{fig:kohler}). PCCA+ has an intrinsic regularization mechanism by which it is able to cluster complex spaces neatly. 
\subsubsection{Spatial Abstraction using PCCA+}
The transition structure of the MDP could be arbitrarily complex and hence, to identify abstractions that are well aligned to the state space transition structure, we use PCCA+ to abstract the MDP in place of other clustering algorithms. This is inpired by \cite{weber2004perron} who use PCCA+ to detect the conformal states of a dynamical system by operating on the transition structure. On providing the graph (transition structure) of the MDP to PCCA+, we derive the abstraction of the MDP where the simplex vertices identified by PCCA+ are the {\it abstract states}. We also get a membership function from PCCA+, $\chi$, which defines the degree of membership of each state $s$ to an abstract state $S$. We describe how this information can used to compose options in the next section.

\section{Composing Options from PCCA+}

Given the membership functions and the abstractions discovered by PCCA+, we provide a very elegant way to compose option policies to move from one abstract state to another. The transition to another abstract state is done simply by {\it following the positive gradient (hill climbing) of the membership value to the destination abstract state}. This navigates the agent through states whose membership functions to the target abstract state progressively increases.

Specifically, consider that we have $N$ states $s_1, s_2, s_3, \cdots, s_N$ and $k$ abstract states $S_1, S_2, \cdots, S_k$. Typically, $N \gg k$. Let $\chi_{ij}$ denote the membership of state $s_i$ to the abstract state $S_j$. $s_i$ is said to {\it belong} to abstract state $S_a$, where $a = \textrm{argmax}_{j} \chi_{ij}$. Thus, any option between abstract states $S_i$ and $S_j$ will start at a state in $S_i$ and end at a state in $S_j$. The connectivity information between two abstract states $(S_i, S_j)$ is given by the $(i,j)^{\textrm{th}}$ entry of $\chi^T L \chi$ where $L$ is the Laplacian corresponding to the MDP transition matrix. The diagonal entries provide the relative connectivity information within a cluster. We generate an option for every pair of connected abstract states. For an option from $S_i$ to $S_j$:

\begin{itemize}
    \item The initiation set $I$ represents states that belong to $S$.
    \item The {\textbf{option policy}} $\mu(s,a)$ that takes the agent from abstract state $S_i$ to $S_j$ is a stochastic gradient function given by:
    $$\mu(s,a) = \alpha(s)\left[\textrm{max}\left((\chi_{S_jS_i}(s,a) - \chi_{S_j}(s)), 0\right)\right]$$ where
    $$\chi_{S_jS_i}(s,a) = \left[\left(\sum_{s'} P(s,a,s')\chi_{S_j}(s')\right) - \chi_{S_j}(s)\right] \ \forall \ s \in S_i$$
    $\alpha(s)$ is a normalisation constant to ensure $\mu$ represents a probability function in the range $[0,1]$.
    \item Finally, {\textbf{termination condition}} $\beta$ is a function which assigns the probability of termination of the current option at a state $s$. It could also be viewed as the probability of a state $s$ being  decision epoch given the current option being executed. For an option taking the agent from abstract state $S_i$ to $S_j$, we define $\beta$ as follows:
    $$\beta(s) = \textrm{min}\left(\frac{\textrm{log}(\chi_{S_{i}}(s))}{\textrm{log}(\chi_{S_{j}}(s))}, 1 \right) \ \forall \ s \in S_i$$
\end{itemize}

We offer an intuitive explanation for the choice of the option policy $\mu$ and option termination condition $\beta$. $\chi_{S_jS_i}(s,a)$ is the expected increase in the membership function to abstract state $S_j$ on taking action $a$ at state $s$, while $\chi_{S_j}(s)$ is the membership function to $S_j$ at current state $s$. The probability of an action $a$ involved in taking the agent from state $s \in S_i$ to an abstract state $S_j$ must proportional to the expected increase in the membership function to the abstract state $S_j$ if positive and $0$ if the expected increase is negative. The choice for the termination condition also has a simple heuristic: When transitioning from $S_i$ to $S_j$, the agent would encounter bottleneck state(s) $s*$ which would approximately satisfy $\chi_{S_i}(s) = \chi_{S_j}(s)$, and hence be suitable for the termination of the option. Till then, the membership value of $S_i$ would be higher than $S_j$ (with the difference gradually reducing along the option trajectory), and hence the probability to terminate would be proportional to how much of a {\it bottleneck} the current state $s$ is. Even though we do not look for {\it bottleneck states} directly in our approach unlike \cite{mcgovern2001automatic}, the termination condition we propose naturally captures transitioning through {\it bottleneck} states.

\section{Online Agent}

The previous section discussed how to discover options given that abstractions are obtained from the transition structure of the MDP. However, an online agent has no prior knowledge of the model and has to learn these abstractions from scratch. We present Algorithms 2 and 3 respectively for an online agent to perform Option Discovery using Spatio-Temporal Clustering (ODSTC) for two cases: i) Small state space task ii) Large State Space Task (where it is infeasible to do model estimation on the original state space).

\begin{algorithm}
\caption{ODSTC Online Agent Pipeline - Small}
\begin{algorithmic}[1]
\WHILE {Not Converged}
    \STATE Sample trajectories using current behavioral policy
    \STATE Estimate model from sample trajectories 
    \STATE Operate PCCA+ on estimated model to derive abstract states and memberships
    \STATE Discover options from the abstract states and memberships
    \STATE Augment agent with new options 
    \STATE Update value functions and behavioral policy using SMDP Q Learning
\ENDWHILE
\end{algorithmic}
\end{algorithm}

\begin{algorithm}
\caption{ODSTC Online Agent Pipeline - Large}
\begin{algorithmic}[1]
\WHILE {Not Converged}
    \STATE Sample trajectories using current behavioral policy
    \STATE Run the trajectory data points through an Action Conditional Video Pred Network
    \STATE Use the learned representation to aggregate states into {\it microstates} through K-Means
    \STATE Estimate model on {\it microstate} space from sample trajectories 
    \STATE Operate PCCA+ on estimated model to derive abstract {\it macrostates} and memberships
    \STATE Discover options from the abstract {\it macrostates} and memberships
    \STATE Augment agent with new options 
    \STATE Update value functions and behavioral policy using Intra-Option Q Learning with DQN
\ENDWHILE
\end{algorithmic}
\end{algorithm}

Here are two caveats in Algorithm 3:
\begin{itemize}
    \item It might be too expensive to perform PCCA+ for large scale problems after every episode and hence we could update the skills or options from PCCA+ after a fixed number of episodes depending on the problem complexity.
    \item Though for small scale problems, the first initial exploration could come from random policies, this wouldn't work for large problems like Seaquest, where an informed exploration with a partially trained Deep Q Network (DQN) \cite{mnih2015human} is necessary to ensure sufficient portion of the state space is seen for learning a good enough aggregate space for PCCA+ to operate on.
\end{itemize}

\subsection{Model estimation and incorporation of reward structure}
For every sampled trajectory, we maintain transition counts $\phi_{ss'}^{a}$ of the number of times the transition $s \xrightarrow{a} s'$, starting from $\phi_{0}$. These transition counts are used to populate the local adjacency matrix $D$ and the transition count model as: $D_{\textrm{posterior}}(s,s')=D_{\textrm{prior}}(s,s')+ \sum_a \phi^{a}_{ss'}$, $U_{\textrm{posterior}}(s,a,s') = U_{\textrm{prior}}(s,a,s') + \phi_{ss'}^a$, after every sampled trajectory. PCCA+ is used to find suitable spatial abstractions and subtask options where the transition probabilities in the hill-climbing step are calculated as $P(s,a,s') = \frac{U(s,a,s'}{\sum_{s'} U(s,a,s')}$. We then use an SMDP value learning algorithm to find the optimal policy over the constructed options.

The underlying transition structure $\phi_{ss'}^{a}$ encodes only the topological properties of the state space. There are other kinds of structures in the environments which we would like an autonomous agent to discover. For example, a monkey climbing a tree to pick up a fruit abstracts the task as to reach the branch of the tree nearest to the fruit. These structures are functional in nature, depending on the functional properties of the task. We use reward counts for each transition to encode the functional properties in the transition structure. The idea is that the spatio-temporal abstractions should {\it degenerate} in places where there is a spike in the reward distribution, so that our agent interprets a state of high reward (as goal states typically are) as a different abstract state, and naturally composes options that would lead it to that state. Let $R_{ss'}^a$ be the reward in performing a transition $s \xrightarrow{a} s'$. We modify the local adjacency matrix $D$'s posterior update as: $D_{\textrm{posterior}}=D_{\textrm{prior}}(s,s')+ \sum_a \phi^{a}_{ss'}e^{-v|R_{ss'}^{a}|}$ where $v$ is the regularization constant that balances the relative weight of the underlying reward and transition distribution. We use an exponential weighting for rewards to ensure that the adjacency function has a very low value at the spike points where we want to abstraction to degenerate. It also returns a value of $1$ for zero rewards, preserving the original transition structure and allows easy tuning of relative weights ($v$).

\section{Experiments}

\begin{figure}
\centering

\subfloat[3-Room domain]{
        \label{fig:3room}
        \includegraphics[width=0.165\textwidth]{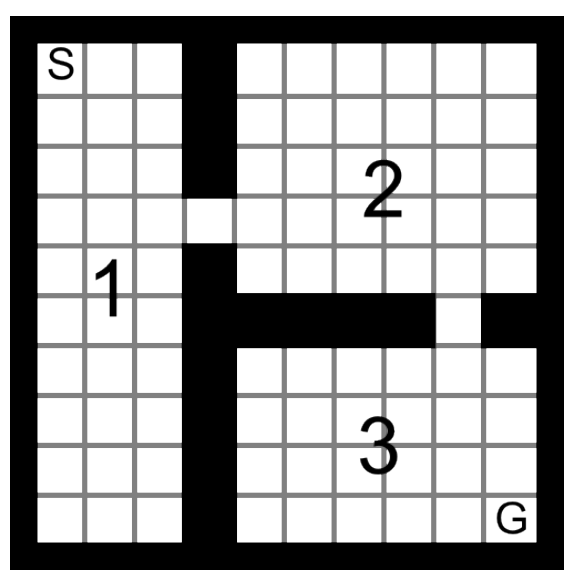}}
\subfloat[Membership Function]{
        \label{fig:flatmembership}
        \includegraphics[width=0.165\textwidth]{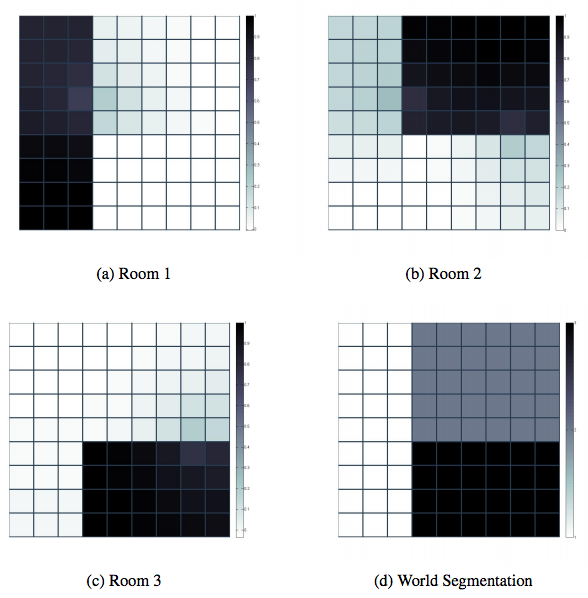}} 

\subfloat[Effect of Reward Structure Incorporation]{
        \label{fig:rewardimpact}
        \includegraphics[width=0.45\textwidth]{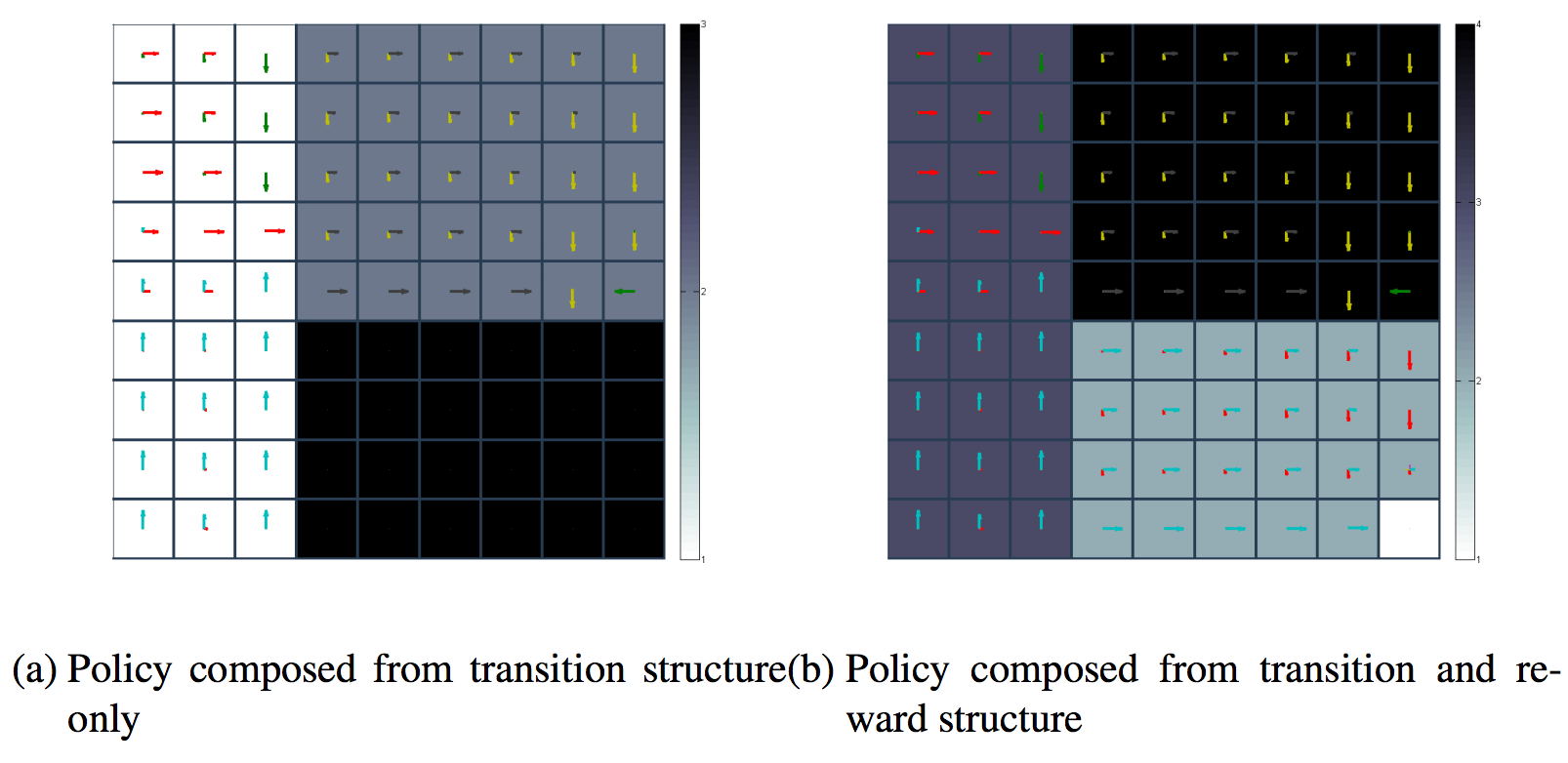}} 
\caption{3 Room Domain Visualization}
\label{fig:3roomfigures}

\end{figure}
{\textbf{3 Room Domain:}} Let us consider a simple 3 room domain as shown. The world is composed of rooms separated by walls as shown in Fig \ref{fig:3room}. For the current task, let us assume that the goal of the agent is to start from the tile marked $S$ and reach the goal tile marked $G$. The agent can either move a step in the direction towards north, east, west or south. PCCA+ discovers three abstract states without incorporating the reward structure, each corresponding to one room in the 3-room world. On incorporation of the reward structure, there would be $4$ abstract states, with the additional one corresponding to the goal state alone. The doorway could be seen as bottleneck states at which the membership functions lead to termination conditions of options exiting one abstract state (current room) and entering another abstract (destination room). The navigation task could simply be abstracted as exiting the current room by moving to the doorway and entering the correct room to reach the goal state in it. We visualize the membership function identified by PCCA+ on the 3-room world in Fig \ref{fig:flatmembership} without reward structure incorporation. It is clear that each abstract state can be identified with one of the three rooms in the world and the MDP transition structure has been neatly segmented by PCCA+. Fig \ref{fig:rewardimpact} shows the MDP segmentation and the option policies discovered without and with the reward structure incorporated. For the first case (left), without the reward structure, there are only $3$ abstract states corresponding to each room and hence the option policies discovered would be navigation from one room to another. The figure shows stochastic option policies to navigate to room $3$ from the other two rooms. In the second case (right), with the reward (goal) considered in the transition structure, $4$ abstract states are identified, with the new one lonely being the goal state in the bottom right corner as indicated. The figure shows the (stochastic) option policies in getting to room $3$ from rooms $1$ and $2$, and in getting from the group of granular states other than the goal state in room $3$ to goal state. We hence clearly see the advantage of discovering these {\it structural} (from transition structure) and {\it functional} (from reward structure) abstractions since the solving the task now is as simple as planning to move from one abstract state to another instead of trying to learn the optimal policy at every granular state in the MDP.

\begin{figure}
\centering
\includegraphics[width=0.4\textwidth]{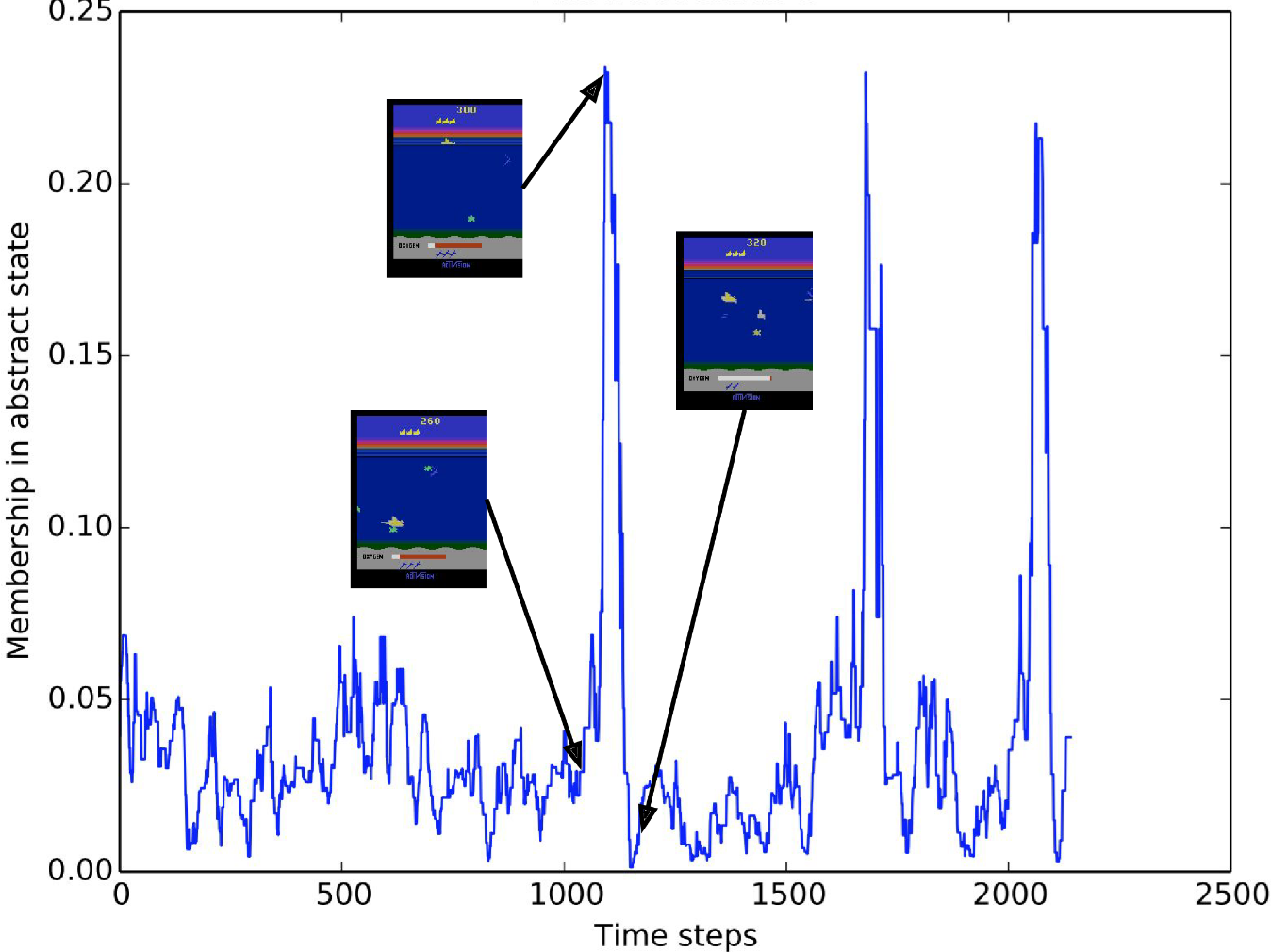}
\caption{Seaquest Membership Function Visualization}
\label{fig:seaquest}
\end{figure}

{\textbf{Seaquest:}} Atari 2600 games have been one of the widely used benchmarks for reinforcement learning algorithms on large state spaces \cite{mnih2015human}, \cite{mnih2016asynchronous}, etc. Due to infeasibility in applying the PCCA+ framework directly on the exponential state space, we perform a {\it state space aggregation} by using K-Means clustering. However, aggregating directly on the input pixel space ($210\times160\times3$) in Seaquest is infeasible and no latent concepts (spatial and temporal) are captured in such a space. We therefore adopt the Action Conditional Video Prediction Network of \cite{oh2015action} for learning a latent space that captures the spatio-temporal concepts. The overall pipeline is described in Algorithm 3. Fig \ref{fig:seaquest} helps to appreciate the semantics of an example option generated in Seaquest. We plot the agent's membership in the abstract state corresponding to particular option for a randomly chosen episode's trajectory. The skill learnt here corresponds to resurfacing to replenish oxygen in Seaquest.

\section{Conclusion}

In this paper, we have presented a new framework for automated skill acquisition in reinforcement learning that builds an abstraction over the MDP well aligned to the transition structure and discovers skills or options that enable movement across these abstract states. The framework works without an exact model of the MDP and the option policies are obtained in a very elegant way by performing hill climbing on the membership function of the abstract states. These options can also be efficiently reused across multiple tasks sharing a common transition structure. We also discuss an approach to scaling the pipeline to a larger state spaces by performing state aggregation through clustering on the representation discovered by an unsupervised model prediction network. 





\bibliographystyle{aaai}
\bibliography{aaai}

\end{document}